\documentclass[10pt,twocolumn,letterpaper]{article}

\usepackage[pagenumbers]{iccv} 

\usepackage[table,dvipsnames]{xcolor}
\usepackage{makecell}
\usepackage{multirow}
\usepackage{colortbl}
\usepackage{amssymb}
\usepackage{fontawesome}
\usepackage{marvosym}
\definecolor{iccvblue}{rgb}{0.21,0.49,0.74}
\usepackage[pagebackref,breaklinks,colorlinks,allcolors=iccvblue]{hyperref}

\definecolor{lora_orange}{RGB}{233, 113, 50}
\definecolor{lora_blue}{RGB}{78, 149, 217}

\definecolor{lightgray}{gray}{0.9}
\definecolor{linecolor}{RGB}{204, 204, 255}
\usepackage[capitalize]{cleveref}
\crefname{section}{Sec.}{Secs.}
\Crefname{section}{Section}{Sections}
\Crefname{table}{Table}{Tables}
\crefname{table}{Tab.}{Tabs.}
\definecolor{mygray}{gray}{.9}
\definecolor{mygray1}{gray}{.92}
\definecolor{evaunit01green}{RGB}{82,208,83}
\definecolor{lowred}{RGB}{238,18,137}

\definecolor{lowerred}{RGB}{255,110,180}

\definecolor{defaultcolor}{RGB}{12,127,17}

\def\paperID{3167} 
\def\confName{ICCV}
\def\confYear{2025}

\title{SAM4D: Segment Anything in Camera and LiDAR Streams}

\author{ 
  Jianyun Xu$^{1,\dag}$ \ \ Song Wang$^{1,2,\dag}$ \ \ Ziqian Ni$^{1,\dag}$ \ \ Chunyong Hu$^1$ \ \ Sheng Yang$^{1,}$\textsuperscript{\Letter} \ \ Jianke Zhu$^2$  \ \ Qiang Li$^1$  \\[0.1cm]
  {
        $^1$Unmanned Vehicle Dept., CaiNiao Inc., Alibaba Group \ \ \
        $^2$Zhejiang University } \\  [0.1cm]
        \faGithubAlt~\textbf{Project Page:} \href{https://sam4d-project.github.io}{\texttt{SAM4D-Project.github.io}}
}

\begin{document}

\twocolumn[{
\maketitle
\centering
\captionsetup{type=figure}
\includegraphics[width=0.99\textwidth]{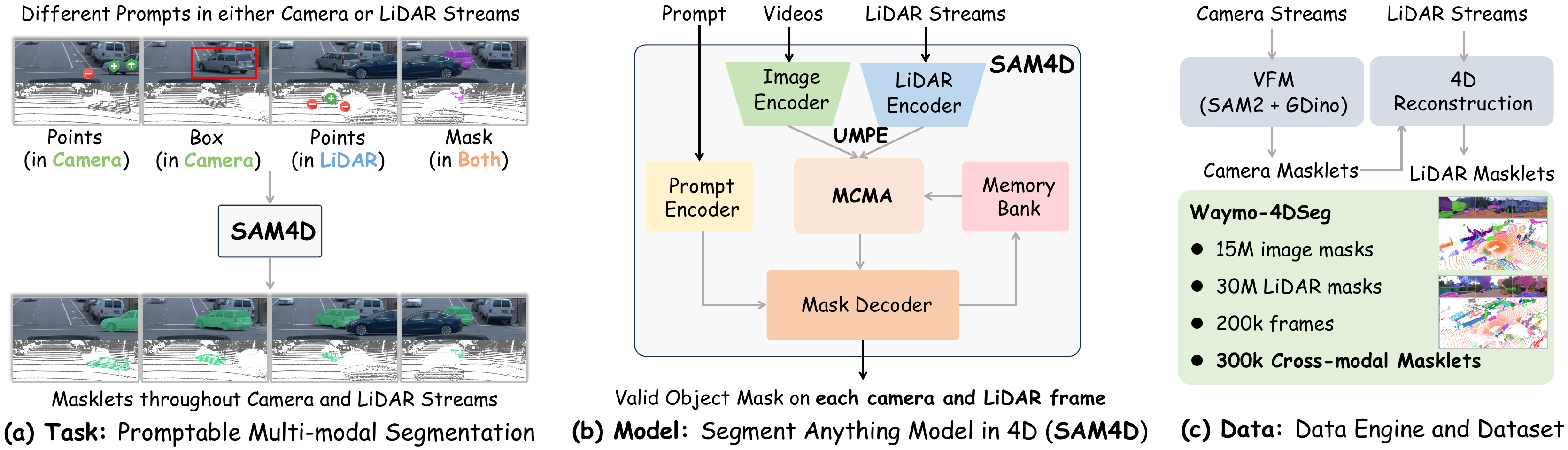}
\vspace{-1mm}
\captionof{figure}{We aim to build a foundation model for 4D segmentation by introducing three interconnected components: (a) a promptable multi-modal segmentation \textbf{task}, extending segmentation to both camera and LiDAR streams; (b) a segmentation \textbf{model} (SAM4D) that enables 2D-3D joint segmentation with cross-modal prompting and temporal alignment; and (c) an automatic \textbf{data} engine for constructing Waymo-4DSeg, a large-scale dataset with over $300$k camera-LiDAR associated masklets, providing pseudo labels for SAM4D training.}
\label{fig:teaser}
\vspace{5mm}
}]

\maketitle
{\renewcommand\thefootnote{}\footnotetext{$\dag$: Equal contribution. \Letter: Corresponding author.}}

\begin{abstract}
We present SAM4D, a multi-modal and temporal foundation model designed for promptable segmentation across camera and LiDAR streams. Unified Multi-modal Positional Encoding (UMPE) is introduced to align camera and LiDAR features in a shared 3D space, enabling seamless cross-modal prompting and interaction. Additionally, we propose Motion-aware Cross-modal Memory Attention (MCMA), which leverages ego-motion compensation to enhance temporal consistency and long-horizon feature retrieval, ensuring robust segmentation across dynamically changing autonomous driving scenes. To avoid annotation bottlenecks, we develop a multi-modal automated data engine that synergizes VFM-driven video masklets, spatiotemporal 4D reconstruction, and cross-modal masklet fusion. This framework generates camera-LiDAR aligned pseudo-labels at a speed orders of magnitude faster than human annotation while preserving VFM-derived semantic fidelity in point cloud representations. We conduct extensive experiments on the constructed Waymo-4DSeg, which demonstrate the powerful cross-modal segmentation ability and great potential in data annotation of proposed SAM4D.
\end{abstract}
\section{Introduction}
\label{sec:intro}

Segment Anything Model (SAM)~\cite{kirillov2023segment} has emerged as a foundation model for promptable visual segmentation, demonstrating strong generalization in diverse image domains through user-defined prompts such as points, boxes, and masks. Building on this, SAM2~\cite{ravi2024sam} extends segmentation to videos by incorporating a data engine for large-scale video annotation and a streaming memory mechanism for real-time processing. 
These advances highlight the potential of promptable segmentation in various downstream tasks~\cite{ma2024segment, mazurowski2023segment, chen2024rsprompter, tang2023can, chen2024sam2}. 
However, existing methods remain limited to the image and video domains without considering other sensor modalities crucial for safety-critical applications such as autonomous driving.

Achieving higher levels of autonomy in driving systems requires robust multi-modal perception~\cite{liu2023bevfusion, yan2023cross, zhang2024sparselif}, where cameras and LiDAR synergistically compensate for each other's limitations, particularly in challenging conditions such as low visibility or poor lighting~\cite{xie2023robobev,dong2023benchmarking}. Although active depth sensing of LiDAR provides precise geometric priors and enables direct temporal feature association, existing segmentation models for LiDAR perception~\cite{zhou2025pointsam, ovsep2024better, liu2023segment} remain largely frame-centric. 
To the best of our knowledge, no prior work has systematically leveraged cross-modal spatial consistency across synchronized LiDAR scans and camera streams for both 2D and 3D segmentation. These oversights limit the efficacy of joint image and LiDAR annotation, where temporal and cross-modal cues is critical to resolve ambiguities and insufficient observations. 

To reduce annotation costs and improve multi-modal segmentation efficiency, we introduce the Promptable Multi-modal Segmentation (PMS) \textbf{task}, which enables segmentation across camera and LiDAR sequences based on prompts (\textit{e.g.}, points, boxes, or masks) from both modalities.  Furthermore, cross-modal prompting is introduced, allowing a query in one modality (\textit{e.g.}, an image prompt) to guide segmentation in another (\textit{e.g.}, LiDAR).

Based on PMS task, we propose SAM4D, \textit{the first promptable multi-modal segmentation} \textbf{model} for camera and LiDAR streams, unifying multi-modal and temporal segmentation within a single framework.
Specifically, SAM4D is built upon a multi-modal transformer architecture, integrating Unified Multi-modal Positional Encoding (UMPE) for spatial alignment and Motion-aware Cross-modal Memory Attention (MCMA) for temporal consistency. With UMPE, SAM4D explicitly fuses image and LiDAR features in a shared 3D space, enabling seamless cross-modal prompting and interaction through unified positional encoding. Additionally, MCMA incorporates ego-motion compensation, ensuring accurate temporal feature alignment and enhancing long-horizon object tracking in dynamic environments. 
By integrating multi-modal feature fusion, temporal reasoning, and cross-modal interaction, SAM4D is expected to significantly reduce manual labeling efforts while ensuring robust and temporally consistent segmentation in driving scenarios.

To train SAM4D, we construct Waymo-4DSeg, a large-scale multi-modal segmentation \textbf{dataset} based on the Waymo Open Dataset~\cite{sun2020scalability}, designed to provide high-quality, temporally consistent pseudo-ground truth. 
Our proposed multi-modal data engine enhances 2D-3D joint annotation by integrating vision foundation model (VFM)-based video masklet generation, 4D reconstruction for LiDAR pseudo-labeling, and cross-modal label fusion. 
In contrast to previous methods~\cite{zhou2025pointsam, ovsep2024better} that focus on independent frame annotations, our approach uses a sequence-level propagation strategy, ensuring temporal consistency and cross-modal coherence. This significantly improves annotation efficiency and accuracy, making Waymo-4DSeg a key benchmark for training and evaluating promptable, multi-modal, and temporally aware segmentation models for autonomous driving. 
Extensive experiments are conducted with Waymo-4DSeg and unseen dataset under different challenging settings, which demonstrate the strong performance and generalizability of SAM4D in promotable multi-modal segmentation.

\section{Related Work}

\noindent \textbf{Image and LiDAR Segmentation.}
Segment Anything (SAM) series~\cite{kirillov2023segment, ravi2024sam} introduced a foundation model for image and video segmentation, capable of generating masks based on diverse prompt types while demonstrating strong generalization across various datasets. 
Subsequent work has focused on improving the granularity of SAM segmentation~\cite{ke2023segment} and efficiency~\cite{xiong2024efficientsam, zhang2023faster, zhao2023fast}, optimizing its performance for finer segmentation tasks. In addition, researchers have explored the applicability of SAM in various 2D downstream tasks~\cite{mazurowski2023segment, ma2024segment, chen2024rsprompter, tang2023can} and LiDAR point clouds~\cite{liu2023segment,yang2023sam3d, huang2024segment3d, ovsep2024better}. 
Liu \textit{et al.}~\cite{liu2023segment} proposed distilling 2D segmentation masks from SAM into LiDAR-based networks to enhance 3D segmentation performance. Other methods~\cite{yang2023sam3d, huang2024segment3d} project SAM-generated 2D masks into 3D space for further refinement. 
Approaches such as SAL~\cite{ovsep2024better} and PointSAM~\cite{zhou2025pointsam} attempt to build SAM-like promptable segmentation networks directly on point clouds, designing architectures specifically for 3D data. 
There are also studies~\cite{han2024scale, fradlin2024interactive4d} that explore the accumulation of sequential LiDAR point clouds for interactive segmentation that leverages temporal consistency by aggregating point cloud information over multiple frames. 
However, the above methods remain modality-specific, focusing on either image segmentation or LiDAR segmentation in isolation. In contrast, our work is the first to unify image and LiDAR segmentation within a single framework.

\noindent \textbf{Multi-Modal Perception in Driving.}
Recent works have explored multi-modal fusion strategies for Cameras and LiDAR in different spatial representations to improve detection~\cite{liu2023bevfusion, yan2023cross, zhang2024sparselif, chen2023futr3d, li2022unifying}, segmentation~\cite{shin2022mm, wang2022meta, li2023mseg3d, ni2023robust, wang2023lidar2map, cao2024mopa}, and occupancy prediction~\cite{pan2024co, li2024pmafusion, wang2024occgen, zhang2024fusionocc} in autonomous systems. Bird’s-Eye View (BEV) fusion~\cite{liu2023bevfusion, liang2022bevfusion} has gained popularity, where image and LiDAR features are projected into a unified BEV space to facilitate spatial reasoning. Meanwhile, voxel-based fusion~\cite{li2022unifying, wang2024occgen, zhang2024fusionocc} operates directly in 3D space, where features of multiple sensors are aggregated into structured voxel grids for fine-grained 3D perception. Sparse representation-based fusion~\cite{yan2023cross, zhang2024sparselif, xie2023sparsefusion} has also gained attention for its efficient feature encoding and wider perceptual coverage, making it a promising direction in multimodal perception research.
Despite these advancements, most existing methods output only 3D predictions, lacking exploration of cross-modal interactions and unified 2D-3D segmentation. Our work fills this gap by introducing SAM4D, enabling promptable segmentation across camera and LiDAR streams. 
\begin{figure*}[t]\centering
\includegraphics[width=0.95\linewidth]{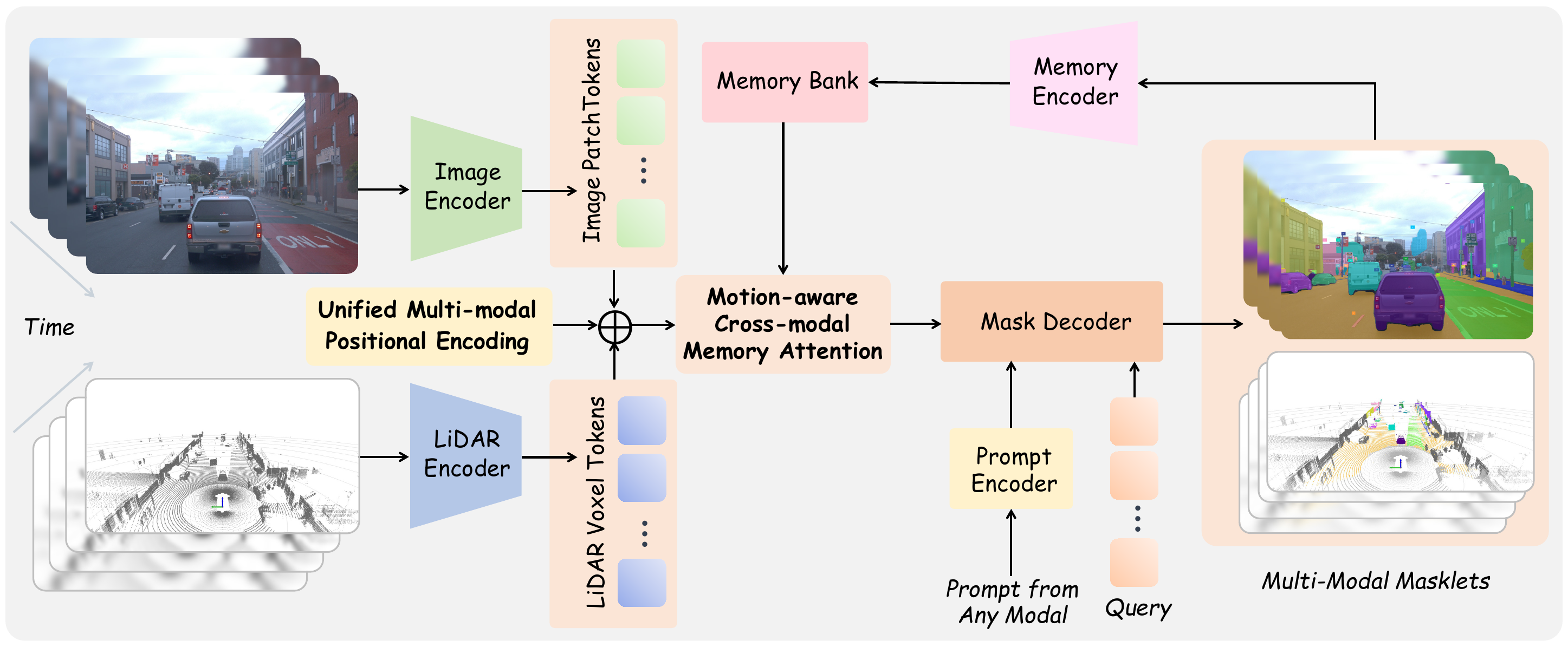}\vspace{-1mm}
\caption{
\textbf{Overview of the Segment Anything Model in 4D (SAM4D) workflow.}
The image and LiDAR encoders generate modality-specific embeddings, which are aligned through the proposed \textbf{Unified Multi-modal Positional Encoding}. The \textbf{Motion-aware Cross-modal Memory Attention} then processes multi-modal and temporal features, incorporating ego-motion for improved feature interaction. Finally, the updated image and LiDAR features are queried efficiently by mask decoder with diverse input prompts from various modalities.
}
\label{fig:model_diagram}
\vspace{-3mm}
\end{figure*}

\noindent \textbf{Lifting Vision Foundation Models for 3D Labeling.}
Recent works have leveraged Vision Foundation Models (VFMs)~\cite{radford2021learning,kirillov2023segment} to enable label-efficient 3D scene understanding by distilling 2D vision priors into 3D representations. 
Methods like CLIP2Scene~\cite{chen2023clip2scene} and OpenScene~\cite{peng2023openscene} transfer CLIP’s vision-language embeddings to 3D semantic segmentation, while approaches such as CLIP-FO3D~\cite{zhang2023clip} and OVO~\cite{tan2023ovo} extend open-vocabulary segmentation to 3D occupancy and point cloud learning. More recent works~\cite{vobecky2024pop, boeder2024langocc} integrate volume rendering techniques to improve 3D occupancy prediction. 
Multimodal fusion strategies have also been explored, with VLM2Scene~\cite{liao2024vlm2scene} incorporating image, text, and LiDAR representations and VEON~\cite{zheng2024veon} enhancing 3D occupancy prediction through vocabulary-driven alignment. Furthermore, methods~\cite{zeng2023clip2,lu2023see} focus on segmentation of zero-shot point clouds by integrating multimodal visual cues. Although these approaches effectively bridge 2D and 3D feature spaces, they are often limited to individual frames and do not explicitly consider temporal consistency. In contrast, our work introduces a 4D data engine that propagates high-quality labels across entire sequences through temporal reconstruction.
\section{Promptable Multi-modal Segmentation}
\label{sec:task}

The Promptable Multimodal Segmentation (PMS) task is designed to enable \textit{interactive}, \textit{cross-modal}, and \textit{temporal} segmentation across both camera and LiDAR streams. Unlike traditional segmentation tasks that rely on a single modality or frame-by-frame processing, PMS allows prompts in either 2D (images) or 3D (LiDAR point clouds) to guide segmentation across the entire sequence. A prompt can be in the form of positive/negative clicks, boxes, or masks that either define a new object or refine an existing segmentation. 
Once a prompt is provided on a specific image frame or LiDAR scan, the model should immediately return valid segmentation masks for both modalities and then propagate the segmentation across the entire sequence, forming masklets that maintain temporal consistency. 
Additionally, PMS allows users to provide additional prompts at any frame in the sequence, refining segmentation across frames and modalities as needed. 

To support PMS, we develop SAM4D, a unified segmentation model capable of processing both videos and LiDAR streams with cross-modal prompting. Additionally, we construct a large-scale dataset based on the Waymo Open Dataset~\cite{sun2020scalability} to provide high-quality pseudo-ground truth annotations for PMS. 
We evaluate SAM4D by simulating interactive multi-modal segmentation scenarios, assessing its ability to segment and track objects across frames and sensor modalities. 
\section{SAM4D Model}
\subsection{Overview}
SAM4D extends SAM2~\cite{ravi2024sam} beyond video segmentation to the multimodal domain, addressing the challenges of cross-modal and long-term object segmentation in autonomous driving scenarios.
Unified multi-modal positional encoding is proposed to enable the multi-modal feature and prompt interaction.
To enhance the ability for long-term object segmentation, we take into account of the ego-motion and design motion-aware cross-modal memory attention.
The overview of our proposed SAM4D is illustrated in Fig.~\ref{fig:model_diagram}.

\subsection{Multi-modal Segmentation Framework}

\noindent \textbf{Multi-modal Feature Embedding.} 
In the video branch, we follow SAM2~\cite{ravi2024sam} and adopt Hiera~\cite{ryali2023hiera, bolya2023window} with SA-V~\cite{ravi2024sam} pre-training to embed each image frame into unconditional patch tokens. 
In the LiDAR branch, MinkUNet~\cite{choy20194d}, implemented with TorchSparse~\cite{tang2022torchsparse,tangandyang2023torchsparse++}, is utilized to encode sparse point clouds into voxel-level tokens.
Throughout the entire interaction process, the image and LiDAR encoder run only once to reduce computational overhead, enabling efficient processing of long-horizon video sequences.

\noindent \textbf{Motion-aware Cross-modal Memory Attention.} 
Our memory attention refines feature representations by integrating cross-modal features and previous frame features in memory (see below) to ensure cross-modal and temporal alignment, which is a core component of our method.
Unlike SAM2, SAM4D lifts image patches into 3D space via depth estimation, allowing unified positional encoding for image patch tokens and LiDAR voxel tokens (see \cref{sec:mm-pe}). 
Furthermore, ego-motion is also embedded in cross-attention with past features and predictions to enable long-term temporal consistency (see \cref{sec:mcma}).

\noindent \textbf{Prompt Encoder and Mask Decoder.} The prompt encoder supports different input prompts from both the image and LiDAR inputs to define the spatial extent and position of the target. Sparse prompts, such as points and boxes, are represented by positional encoding (see \cref{sec:mm-pe}) summed with learnable embeddings for each type of prompt, while mask prompts are embedded using convolutions for images and sparse convolutions for LiDAR. 
The mask decoder processes the prompts from both modalities along with the image and LiDAR features updated by memory attention, simultaneously predicting 2D and 3D segmentation masks. 

\noindent \textbf{Memory Encoder and Memory Bank.} 
The memory encoder processes both 2D and 3D segmentation masks separately, utilizing convolutions for the image and sparse convolutions for LiDAR to downsample the output. 
The downsampled masks are then summed element-wise with the initial embeddings from the image and LiDAR encoders, respectively. A lightweight convolutional layer is then applied to fuse the information, generating the final representation in the memory. The memory bank maintains a FIFO queue to store past object features, with up to $N$ unprompted frames being retained. Additionally, a separate FIFO queue stores $M$ prompted frames to preserve keyframes with explicit user input. We store object pointers computed from mask decoder tokens for both the image and LiDAR domains, capturing high-level semantic information of segmented objects and participating in memory attention.

\begin{figure}[t]\centering
\includegraphics[width=0.95\linewidth]{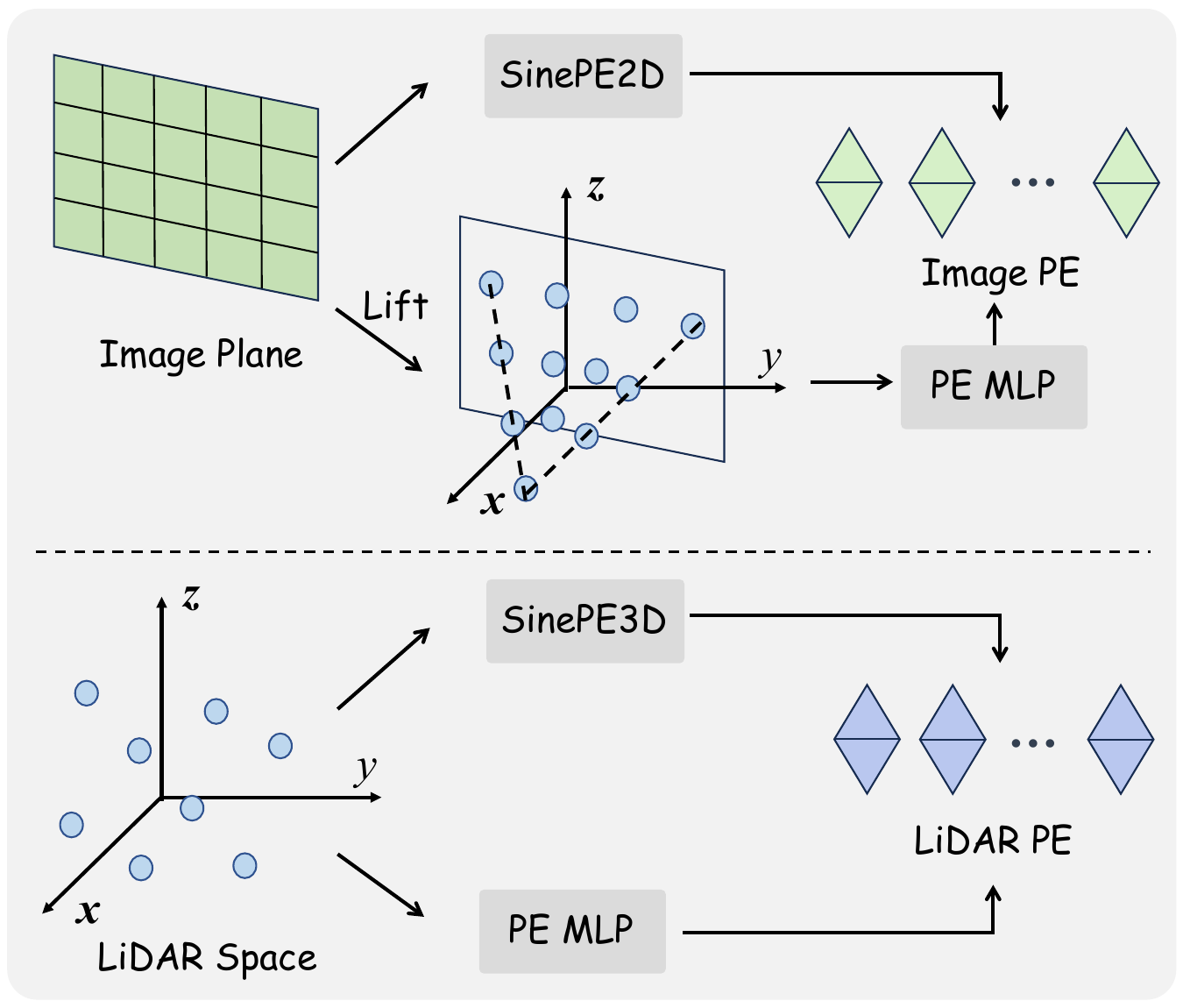}\vspace{-3mm}
\caption{Illustration of the proposed \textbf{Unified Multi-modal Positional Encoding}.}
\label{fig:um_pe}
\vspace{-5mm}
\end{figure}

\subsection{Unified Multi-modal Positional Encoding}
\label{sec:mm-pe}
To ensure consistent spatial representation for both image and LiDAR modalities, we carefully design a Unified Multi-modal Positional Encoding (UMPE) scheme. 
As shown in \cref{fig:um_pe}, this encoding unifies the 2D and 3D features in a shared spatial space, allowing cross-modal interactions while preserving the intrinsic structure of each modality. 
UMPE consists of two complementary components: (i) a modality-specific positional prior, which encodes features in their native spaces, and (ii) a shared 3D representation, which aligns both modalities in a common spatial domain.

\noindent \textbf{Positional Encoding for Images.}
For a pixel  $\mathbf{p} = (u, v)$ in an image feature, we first assign a 2D sinusoidal positional encoding ($\texttt{SinPE2D}$):
\begin{equation}
\mathcal{P}_{\text{img\_sin}} = \texttt{SinPE2D}(u, v),
\end{equation}
which preserves spatial structure in the image plane.
To align image features with LiDAR spatial representations, we estimate a set of depths  $D(u,v)$  for each pixel and lift it into 3D space, similar to Lift-Splat-Shoot~\cite{philion2020lift}:
\begin{equation}
    \mathbf{x}_{\text{img}} = T_{c}^{l}K^{-1} [u*D(u,v), v*D(u,v), D(u,v), 1]^T,
\end{equation}
where $K \in R^{4\times4}$ is the intrinsic matrix of the camera, and $T_{c}^{l} \in R^{4\times4}$ is the transformation matrix from the camera coordinate to the LiDAR one. This process converts the image into a pseudo-point cloud $\mathbf{x}_{\text{img}}$. We then apply an MLP-based 3D positional encoding:
\begin{equation}
\mathcal{P}_{\text{img\_mlp}} = \texttt{MLP}(\mathbf{x}_{\text{img}}).
\end{equation}
The final position encoding $\mathcal{P}_{\text{img}}$ is composed of $\mathcal{P}_{\text{img\_sin}}$ and $\mathcal{P}_{\text{img\_mlp}}$, and these two parts ensure that the image features are represented in the same spatial domain as LiDAR while maintaining the original view-based structure.

\noindent \textbf{Positional Encoding for LiDAR.}
For a LiDAR point at  $\mathbf{x}_{\text{LiDAR}} = (x, y, z)$ , we follow a similar two-stage encoding to obtain $\mathcal{P}_{\text{LiDAR}}$. First, a 3D sinusoidal positional encoding is applied:
\begin{equation}
\mathcal{P}_{\text{LiDAR\_sin}}(x, y, z) = \texttt{SinPE3D}(x, y, z)
\end{equation}
which encodes the spatial structure of the point cloud.
To ensure consistency with image features lifted into 3D, we utilize the same MLP-based transformation:
\begin{equation}
\mathcal{P}_{\text{LiDAR\_mlp}} = \texttt{MLP}(\mathbf{x}_{\text{LiDAR}}).
\end{equation}
For the convenience, we define a symbol $\Phi$ to represent the positional encoding of these two stages for both modalities.

For sparse prompts incluing points or bounding boxes from the image or LiDAR, we apply the same dual-stage positional encoding as used for dense features. The encoded prompts from both modalities are concatenated before being fed into the mask decoder, where missing prompts from one modality are replaced with an empty placeholder. The sparse prompt embeddings are then concatenated to the output tokens, and then applied cross-attention from features updated by the following motion-aware memory attention, enabling the mask decoder to generate both 2D and 3D segmentation masks.
By unifying image and LiDAR positional encodings in a shared 3D space, while preserving modality-specific characteristics, UMPE enables further cross-modal feature fusion and interaction in our framework.

\begin{figure}[t]\centering
\includegraphics[width=0.95\linewidth]{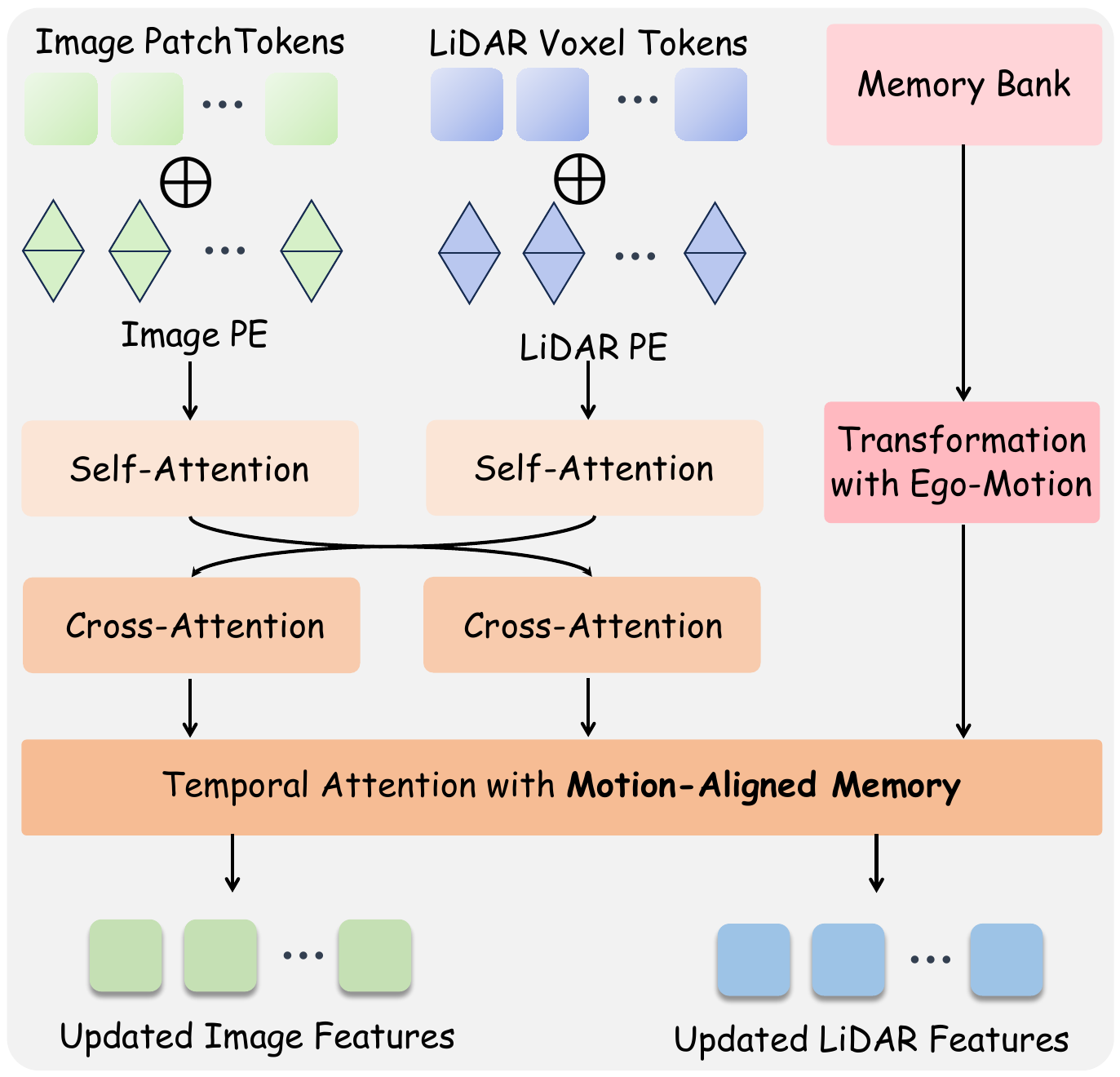}\vspace{-3mm}
\caption{Illustration of the proposed \textbf{Motion-aware Cross-modal Memory Attention}.}
\label{fig:mc_ma}
\vspace{-5mm}
\end{figure}

\subsection{Motion-aware Cross-modal Memory Attention}
\label{sec:mcma}
To enhance multi-modal feature representations while ensuring temporal consistency, we introduce Motion-aware Cross-modal Memory Attention (MCMA). This module integrates self-attention, cross-attention across image and LiDAR modalities, and memory-based temporal attention, as illustrated in \cref{fig:mc_ma}. 
A key distinction from previous approaches~\cite{kirillov2023segment, ravi2024sam} is our incorporation of ego-motion compensation, which aligns the features of the past frame to the current coordinate system, allowing for more accurate feature retrieval and reducing errors in dynamically changing autonomous driving scenes.

\noindent \textbf{Self-Attention for Feature Refinement.}
Given the image features  $\mathcal{F}_{\text{img}}$  and LiDAR features  $\mathcal{F}_{\text{LiDAR}}$  from the respective encoders, along with their positional encodings  $\mathcal{P}_{\text{img}}$  and  $\mathcal{P}_{\text{LiDAR}}$  obtained from Unified Multi-modal Positional Encoding (UMPE), we first apply self-attention within each modality to refine intra-modal feature representations:
\begin{equation}
    \begin{aligned}
        \mathcal{F}_{\text{img}}^{\prime} &= \texttt{SelfAttn}(\mathcal{F}_{\text{img}} + \mathcal{P}_{\text{img}}), \\
        \mathcal{F}_{\text{LiDAR}}^{\prime} &= \texttt{SelfAttn}(\mathcal{F}_{\text{LiDAR}} + \mathcal{P}_{\text{LiDAR}}),
    \end{aligned}
\end{equation}
where \texttt{SelfAttn} represents the self-attention, allowing each token to attend to others within the same modality.

\noindent \textbf{Cross-Attention for Multi-modal Fusion.}
To facilitate interaction between image and LiDAR features, we perform cross-attention, enabling one modality to incorporate information from the other:
\begin{equation}
\begin{aligned}
    \mathcal{F}_{\text{img}}^{\prime\prime} &= \texttt{CrossAttn}(\mathcal{F}_{\text{img}}^{\prime}, \mathcal{F}_{\text{LiDAR}}^{\prime} + \mathcal{P}_{\text{LiDAR}}), \\
    \mathcal{F}_{\text{LiDAR}}^{\prime\prime} &= \texttt{CrossAttn}(\mathcal{F}_{\text{LiDAR}}^{\prime}, \mathcal{F}_{\text{img}}^{\prime} + \mathcal{P}_{\text{img}}),
\end{aligned}
\end{equation}
This step ensures that both modalities share complementary spatial and structural information, enhancing feature expressiveness for segmentation.

\noindent \textbf{Temporal Attention with Motion-Aligned Memory.}
In contrast to SAM2~\cite{ravi2024sam}, which only considers short-term object motion, our method explicitly incorporates \textit{ego-motion compensation} to handle large-scale scene changes in autonomous driving scenarios. 
We maintain a memory bank that stores historical image and LiDAR features  $\mathcal{M}_{\text{img}}$, $\mathcal{M}_{\text{LiDAR}}$  along with their 3D space positions $\mathbf{x}_{\text{img}}$ and $\mathbf{x}_{\text{LiDAR}}$. These features and positions are stored in a FIFO queue, keeping $N$ unprompted frames and $M$ prompted frames for temporal reference.

To correctly align past frame features to the current coordinate frame, we transform stored positions using the ego-motion transformation matrix  $T_{t \leftarrow t^{\prime}}$ , which maps historical frame  $t^{\prime}$  to the current frame $t$:
\begin{equation}
\begin{aligned}
    \mathcal{M}_{\text{img}}^{t \leftarrow t^{\prime}} &= \mathcal{M}_{\text{img}}^{t^{\prime}} + \Phi_{\text{img}}(T_{t \leftarrow t^{\prime}}(\mathbf{x}_{\text{img}})), \\
    \mathcal{M}_{\text{LiDAR}}^{t \leftarrow t^{\prime}} &= \mathcal{M}_{\text{LiDAR}}^{t^{\prime}} +  \Phi_{\text{LiDAR}}(T_{t \leftarrow t^{\prime}}(\mathbf{x}_{\text{LiDAR}})),
\end{aligned}
\end{equation}
where  $T_{t \leftarrow t{\prime}} \in SE(3)$  is derived from vehicle odometry, ensuring spatially consistent memory retrieval.  $\Phi_{\text{img}}$ and  $ \Phi_{\text{LiDAR}}$ are the  unified multi-modal positional encoding, consisting of a sinusoidal and a MLP embedding.

\begin{figure*}[t]\centering
\includegraphics[width=0.95\linewidth]{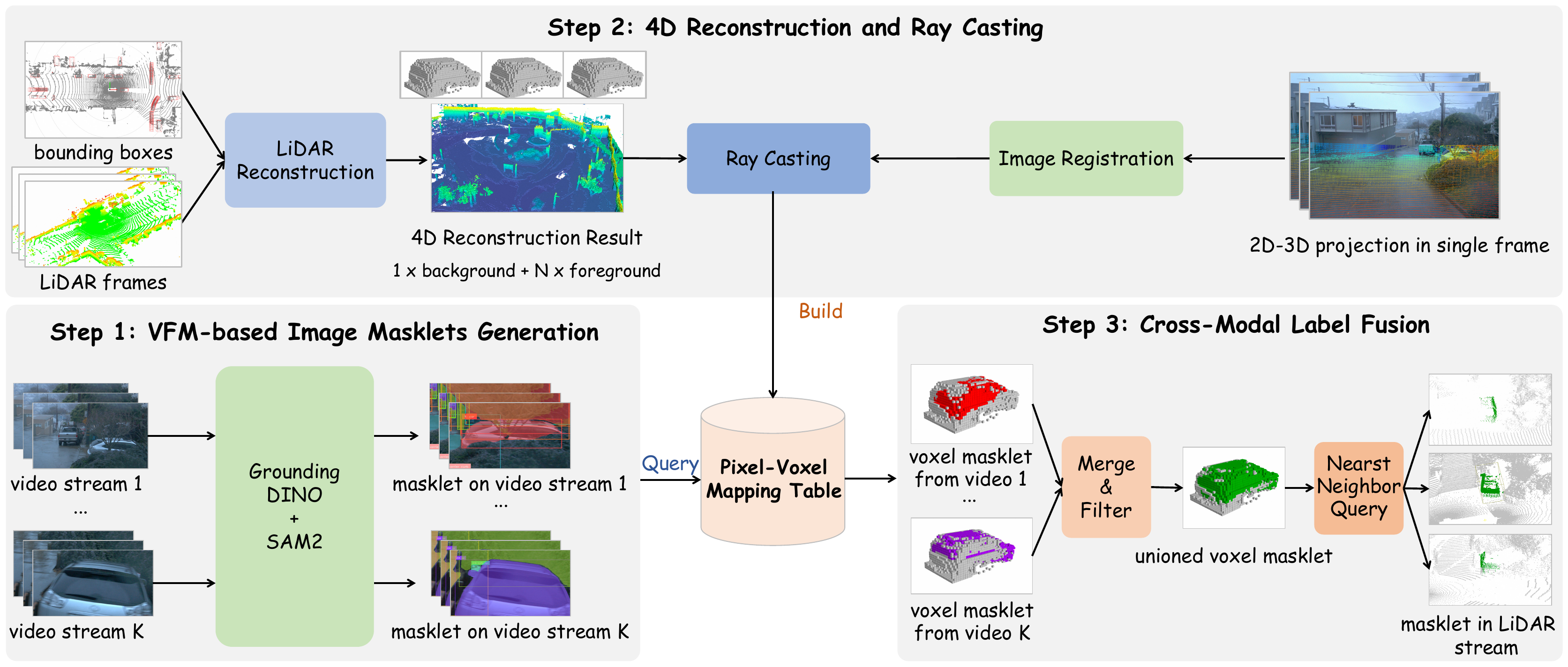}\vspace{-1mm}
\caption{
\textbf{Overview of our data engine}, which is composed of three steps to construct high-quality pseudo labels.  
}
\label{fig:masklet_propagation}
\vspace{-3mm}
\end{figure*}

Similar to SAM2~\cite{ravi2024sam},  we also keep the past object pointers $\mathcal{O}_{\text{img}}^{t^{\prime}}$ and $\mathcal{O}_{\text{LiDAR}}^{t^{\prime}}$ in memory.
Once transformed, previous frame features and object pointers are used in cross-attention to update current features with aligned temporal information:
\begin{equation}
\begin{aligned}
     \mathcal{F}_{\text{img}}^{\text{final}} &= \text{CrossAttn}(\mathcal{F}_{\text{img}}^{\prime\prime}, (\mathcal{M}_{\text{img}}^{t \leftarrow t^{\prime}}, \mathcal{O}_{\text{img}}^{ t^{\prime}})), \\
    \mathcal{F}_{\text{LiDAR}}^{\text{final}} &= \text{CrossAttn}(\mathcal{F}_{\text{LiDAR}}^{\prime\prime}, (\mathcal{M}_{\text{LiDAR}}^{t \leftarrow t^{\prime}}, \mathcal{O}_{\text{LiDAR}}^{t^{\prime}})).
\end{aligned}
\end{equation}
By incorporating motion-aware memory alignment, MCMA significantly improves feature consistency across frames, reducing errors in object correspondence caused by large-scale scene changes. This enables SAM4D to perform robust cross-modal and temporal segmentation in dynamic real-world environments.

\subsection{Training}
The SAM4D model is jointly trained on camera and LiDAR sequences with simulated interactive prompting across modalities, following the strategy introduced in the SAM series~\cite{kirillov2023segment, ravi2024sam}. 
Identical loss functions are applied to both image and LiDAR predictions to enforce cross-modal consistency. Additional training details are provided in the supplementary material.

\section{Data}
To the best of our knowledge, there is currently no dataset that simultaneously supports both 2D and 3D segmentation while ensuring instance consistency over time. 
To quickly establish and expand the training dataset at a low cost, we have carefully designed a multi-modal automatic data engine (\cref{sec:data_engine}) to obtain high-quality pseudo-ground-truth data as much as possible.
Using this data engine, we construct the Waymo-4DSeg dataset (\cref{sec:dataset}) based on the Waymo Open Dataset~\cite{sun2020scalability}, providing a large-scale benchmark for multimodal and temporal segmentation.

\subsection{Data Engine}
\label{sec:data_engine}

Our data engine, as shown in Fig.~\ref{fig:masklet_propagation}, consists of three steps. In Step 1, we leverage vision foundation models (VFM) to generate initial annotations for camera-captured image sequences. 
Given an image sequence of length $T$, we select keyframes at intervals of $K$ frames. Starting from the first frame, we adopt Grounding-DINO~\cite{liu2024grounding, ren2024grounding}, an open-vocabulary detector, and promptable SAM~\cite{kirillov2023segment, ren2024grounded} to obtain detections and segmentation masks for common objects in autonomous driving scenes.
The generated keyframe masks serve as prompts for SAM2~\cite{ravi2024sam}, which propagates the segmentation forward to the next keyframe, producing masklets for the intermediate frames. 

In Step 2, we utilize LiDAR frames and pre-annotated 3D bounding boxes of foreground objects to construct a 4D voxel-based reconstruction, serving as an intermediary between image data and LiDAR frames. This 4D reconstruction consists of a single background component and multiple foreground components, each defined in the body coordinate of the object. We also perform exhaustive ray casting from the center of each image toward the voxels to establish a dense pixel-voxel mapping table.

By querying the pixel-voxel mapping table, we can assign the video masklets to the corresponding voxels in Step 3. However, the presence of noise in the labels and masks necessitated the implementation of a filtering step based on a clustering algorithm. We employ the DBSCAN algorithm to cluster voxels according to their BEV positions and selected the main cluster with the highest vote rate while discarding the rest as noise. After filtering, we assessed overlaps between voxel masklets from different videos to merge them into single masklets. Finally, we created a mapping table between points from LiDAR frames and voxels based on their 3D spatial distances, facilitating the transfer of the final voxel masklet to the LiDAR frames. We evaluated the quality of the resulting masklets using cross-modal IoU, which yielded an average score of $0.56$.

\begin{figure*}[t]
\vspace{-10pt}
\centering
\small
\begin{tabular}{cc}
\includegraphics[width=0.47\linewidth]{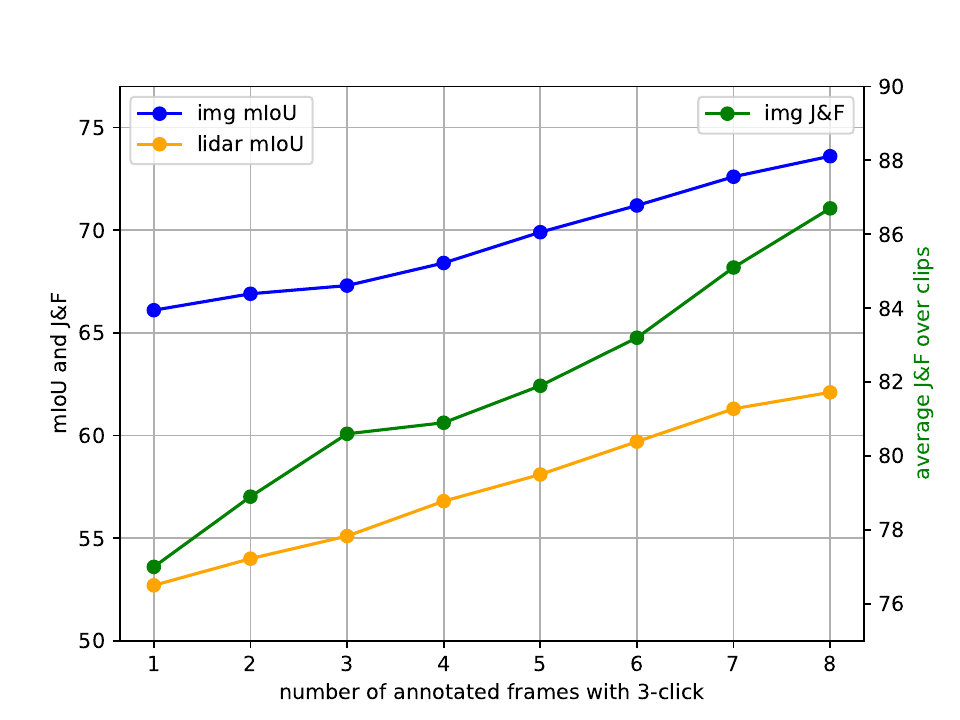} & 
\includegraphics[width=0.47\linewidth]{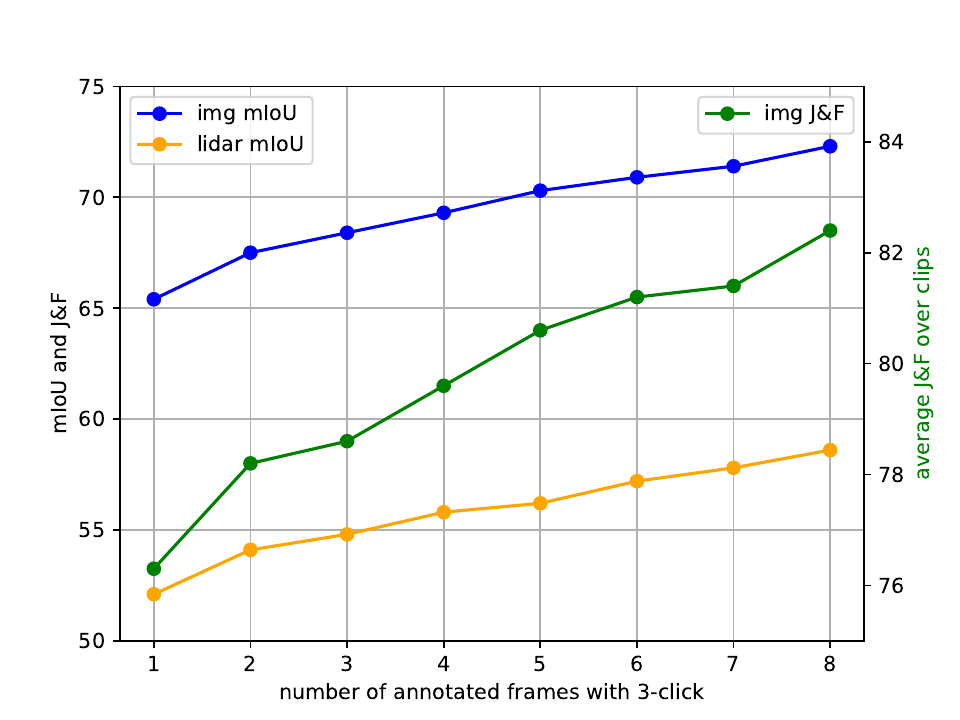} \\
(a) \textit{offline} evaluation ($3$-click) &
(b) \textit{online} evaluation ($3$-click)
\end{tabular}
\vspace{-1mm}
\caption{Performance comparison with different promptable frames in interactive offline and online evaluation settings.}
\vspace{-3mm}
\label{fig:offline_online}
\end{figure*}

\subsection{Constructed Dataset}
\label{sec:dataset}
Our Waymo-4DSeg dataset, derived from the Waymo Open Dataset, follows the original training and validation splits, resulting in $1000$ clips ($798$ for training and $202$ for validation), with about $200$ frames per clip. On average, we generated $300$ masklets for each clip, with each masklet appearing in about $122$ frames. This results in an average of $17$ masks per image and $170$ masks per point cloud. Furthermore, $23.4$\% of the masklets have been observed in at least two different clips.
The semantic categories of our masklets cover nearly all relevant items in autonomous driving scenarios, including dynamic foreground objects (\textit{vehicles}, \textit{pedestrians}), background elements (\textit{buildings}, \textit{trees}) and nearby objects (\textit{curbs}, \textit{lamp posts}, \textit{traffic cones}). The volume of the objects ranges from less than $10$ voxels to over $200$k voxels (with voxel size of $0.1$ meter), occupying an average image area of $1.5$k pixels to over $1$M pixels. More detailed distribution information and visual results are provided in the supplementary material.
\section{Experiments}
\subsection{Setup}
\noindent \textbf{Implementation Details.}
With the Waymo-4DSeg built, we train our SAM4D model with $6$ maximum objects on $16$ NVIDIA A100 GPUs for $36$ epochs. Unless otherwise mentioned, the experimental results in this section are produced in our default setting, using Hiera-S image encoder~\cite{ryali2023hiera} with input image in resolution of $768\times768$ and Mink-34 LiDAR encoder~\cite{choy20194d, tang2022torchsparse} with input LiDAR points voxelized by size of $0.15$m.
More details of the implementation can be found in the supplementary material.

\noindent \textbf{Evaluation Metrics.} 
In the evaluation, the mean Intersection over Union (mIoU) is adopted to assess segmentation performance for both camera and LiDAR in each single frame. 
$\mathcal{J \& F}$ metric~\cite{pont20172017} in video object segmentation is also reported for image sequences. Additionally, we introduce \textit{Number of Mismatched Predictions} (NMP), which quantify the number of instances where a predicted object fails to match the ground truth with an IoU below $0.01$ threshold.  
This metric captures erroneous associations and misalignments, offering information on the robustness of the model to maintain accurate object correspondence between frames. In the practical evaluation, we randomly selected $48$ clips from the validation set.

\subsection{Main Results}

\begin{table}[t]
  \centering
  \caption{
    Performance comparison with different prompts on promptable cross-modal frame segmentation. 
    }
    \vspace{-0.3cm}
    \scriptsize
    \setlength{\tabcolsep}{3.4mm}
    \renewcommand\arraystretch{1.1}
    \begin{tabular}{lcc}
    \toprule
    \textbf{Prompts}  & \textbf{Image mIoU (\%)} $\uparrow$  & \textbf{LiDAR mIoU (\%)} $\uparrow$  \\
    \midrule\midrule
    \rowcolor{mygray1}\multicolumn{3}{c}{\textit{Image-Prioritized Prompting}} \\
    \midrule
    $1$-click  & $68.0$ & $42.3$  \\
    $3$-click & $73.6$ & $\mathbf{53.1}$  \\
    bounding box & $\mathbf{74.7}$ & $47.0$  \\
    \midrule
    \rowcolor{mygray1}\multicolumn{3}{c}{\textit{LiDAR-Prioritized Prompting}} \\
    \midrule
   $1$-click  & $49.6$ & $58.8$  \\
    $3$-click & $\mathbf{64.2}$ & $\mathbf{68.4}$ \\
    bounding box & $46.0$ & $63.9$  \\
    \bottomrule
    \end{tabular}
  \label{tab:cross-prompt}
  \vspace{-3mm}
\end{table}

\noindent \textbf{Promptable Cross-Modal Frame Segmentation.}
For objects captured by both the camera and LiDAR, enabling segmentation in one modality based on a prompt from the other is crucial for improving efficiency in multimodal annotation. 
We evaluated this by selecting objects present in both modalities and providing prompts in one single one (image or LiDAR), then measuring the segmentation IoU in both modalities within a single frame. Prompts include single-point ($1$ click), multi-point ($3$ clicks), bounding box, and mask inputs.
In multiple-click experiments, subsequent clicks are placed on the modality with poorer segmentation after the initial click, simulating human annotation behavior for efficient refinement.
As shown in \cref{tab:cross-prompt}, providing prompts in the image or LiDAR enables the other modality to achieve promising segmentation results, demonstrating the capability of cross-modal prompting in SAM4D.

\noindent \textbf{Promptable Multi-modal Stream Segmentation.}
We further evaluate SAM4D’s stream-level promptable segmentation ability, simulating an interactive annotation process. 
Prompts are given on the first frame where the target appears, with a single-modality prompt if the object is present in one modality, and dual-modality prompts if present in both. 
Similar to SAM2~\cite{ravi2024sam}, experiments are conducted in \textit{offline} and \textit{online} modes. 
In \textit{offline} mode, segmentation is initialized with $3$ point prompts on the first frame, followed by propagation. The frame with the \textit{lowest IoU} is selected for additional prompts, repeating until the prompt limit is reached. 
In \textit{online} mode, segmentation propagates iteratively, adding prompts to frames where IoU falls below $0.75$, until the prompt limit is reached or the sequence ends.
As illustrated in \cref{fig:offline_online}, SAM4D achieves stable segmentation performance in both settings, with continuous improvement as additional prompts are introduced.

\begin{table}[t]
  \centering
  \caption{
    Performance comparison with different prompts on semi-supervised stream object segmentation. 
    }
    \vspace{-0.3cm}
    \scriptsize
    \renewcommand\tabcolsep{0.7pt}
    \renewcommand\arraystretch{1.1}
    \begin{tabular}{lccccc}
    \toprule
    \multirow{2}{*}{\textbf{Prompts}} & \multicolumn{3}{c}{\textbf{Image}} & \multicolumn{2}{c}{\textbf{LiDAR}} \\
    \cmidrule(lr){2-4} \cmidrule(lr){5-6}
    & \textbf{mIoU (\%)} $\uparrow$ & {$\mathbf{\mathcal{J\&F}}$ $\uparrow$ (\%)} & \textbf{NMP} $\downarrow$ & \textbf{mIoU (\%)} $\uparrow$ & \textbf{NMP} $\downarrow$ \\
    \midrule\midrule
    $1$-click  & $61.4$  & $72.2$ & $398$ & $50.1$ & $784$ \\
    $3$-click  & $65.6$ & $76.3$  & $327$ & $52.8$  & $711$ \\
    $5$-click & $67.1$ & $77.7$ & $315$ & $52.6$ & $702$  \\
    bounding box & $64.5$ & $75.4$ & $347$ & $51.3$ & $762$ \\
    \rowcolor{lora_orange!10} ground-truth mask & $\mathbf{69.8}$ & $\mathbf{80.1}$ & $\mathbf{280}$ & $\mathbf{55.7}$ & $\mathbf{582}$ \\
    \bottomrule
    \end{tabular}
  \label{tab:semi}
  \vspace{-5mm}
\end{table}

\noindent \textbf{Semi-Supervised Stream Object Segmentation.}
We extend semi-supervised video object segmentation~\cite{pont20172017, cheng2021modular} to multimodal streams, providing \textit{first frame prompts only} for both image and LiDAR sequences and evaluating segmentation over the full sequence to assess temporal propagation and tracking.
The standard mIoU, $\mathcal{J \& F}$, and NMP are reported for a comprehensive evaluation. As shown in \cref{tab:semi}, mask prompts, which encode richer spatial information, achieve the highest segmentation performance across both modalities, outperforming point and box prompts.

\begin{table}[t]
  \centering
  \caption{
    Performance comparison on nuScenes under semi-
supervised stream object segmentation setting. 
    }
    \vspace{-0.3cm}
    \scriptsize
    \renewcommand\tabcolsep{2.5pt}
    \renewcommand\arraystretch{1.1}
    \begin{tabular}{lccccc}
    \toprule
    \multirow{2}{*}{\textbf{nuScenes}} & \multicolumn{3}{c}{\textbf{Image}} & \multicolumn{2}{c}{\textbf{LiDAR}} \\
    \cmidrule(lr){2-4} \cmidrule(lr){5-6}
    & \textbf{mIoU (\%)} $\uparrow$ & {$\mathbf{\mathcal{J\&F}}$ $\uparrow$ (\%)} & \textbf{NMP} $\downarrow$ & \textbf{mIoU (\%)} $\uparrow$ & \textbf{NMP} $\downarrow$ \\
    \midrule\midrule
    \textit{zero-shot} & $58.4$ & $65.8$ & $36$ & $25.9$ & $117$ \\
    \rowcolor{lora_orange!10} \textit{fine-tuning} & $\mathbf{67.5}$ & $\mathbf{75.4}$ & $\mathbf{22}$ & $\mathbf{44.8}$ &  $\mathbf{70}$\\
    \bottomrule
    \end{tabular}
  \label{tab:nuscenes}
  \vspace{-5mm}
\end{table}
\noindent \textbf{Generalization Experiments on PMS.}
We evaluate SAM4D on unseen nuScenes dataset~\cite{caesar2020nuscenes,li2023lwsis} through \textit{zero-shot} transfer and \textit{fine-tuning} which are detailed in the appendix. 
We adopt the Semi-Supervised Stream Object Segmentation setting, where the mask from the first frame is provided as a prompt to guide the segmentation of subsequent frames.
As shown in \cref{tab:nuscenes}, SAM4D demonstrates strong \textit{zero-shot} segmentation performance, highlighting its cross-modal generalization to unseen driving scenarios. Further \textit{fine-tuning} on nuScenes enhances segmentation quality, demonstrating the model’s ability to adapt and refine predictions in novel environments.

\subsection{Ablations}
\begin{table}[t]
  \centering
  \caption{
    Ablation study on the input modality of SAM4D. 
    }
    \vspace{-0.3cm}
    \scriptsize
    \renewcommand\tabcolsep{1.5pt}
    \renewcommand\arraystretch{1.1}
    \begin{tabular}{lccccc}
    \toprule
    \multirow{2}{*}{\textbf{Input}} & \multicolumn{3}{c}{\textbf{Image}} & \multicolumn{2}{c}{\textbf{LiDAR}} \\
    \cmidrule(lr){2-4} \cmidrule(lr){5-6}
    & \textbf{mIoU (\%)} $\uparrow$ & {$\mathbf{\mathcal{J\&F}}$ $\uparrow$ (\%)} & \textbf{NMP} $\downarrow$ & \textbf{mIoU (\%)} $\uparrow$ & \textbf{NMP} $\downarrow$ \\
    \midrule\midrule
    SAM2+Project & $68.2$ & $79.7$ & $383$ & $32.0$ & - \\ \midrule
    SAM4D-C & $68.6$ & $\textbf{80.4}$ & $301$ & - & - \\
    SAM4D-L & - & - & - & $47.0$ & $799$ \\
    \rowcolor{lora_orange!10} SAM4D & $\mathbf{69.8}$ & $80.1$ & $\mathbf{280}$ & $\mathbf{55.7}$ & $\mathbf{582}$ \\
    \bottomrule
    \end{tabular}
  \label{tab:modal}
  \vspace{-3mm}
\end{table}

We perform ablation studies under the Semi-Supervised Stream Object Segmentation setting to validate the design choices in the SAM4D framework.

\noindent \textbf{Ablation on Input Modality.}
First, we analyze the impact of input modalities by training single-modality variants of SAM4D, where only the image branch (SAM4D-C) or the LiDAR branch (SAM4D-L) is retained, while all other settings are kept the same. 
Additionally, we introduce a baseline (SAM2+Project), which projects SAM2’s video segmentation results onto per-frame point clouds. 
This method is inherently limited by discrepancies in sensor viewpoint, range, and synchronization between camera and LiDAR.
As shown in \cref{tab:modal}, the multimodal SAM4D effectively leverages cross-modal interaction and prompting, achieving significantly better segmentation performance compared to its single-modality counterparts.

\begin{table}[t]
  \centering
  \caption{
    Ablation study on the input resolution of both modalities. 
    }
    \vspace{-0.3cm}
    \scriptsize
    \renewcommand\tabcolsep{1.5pt}
    \renewcommand\arraystretch{1.1}
    \begin{tabular}{lccccc}
    \toprule
    \multirow{2}{*}{\textbf{Resolution}} & \multicolumn{3}{c}{\textbf{Image}} & \multicolumn{2}{c}{\textbf{LiDAR}} \\
    \cmidrule(lr){2-4} \cmidrule(lr){5-6}
    & \textbf{mIoU (\%)} $\uparrow$ & {$\mathbf{\mathcal{J\&F}}$ $\uparrow$ (\%)} & \textbf{NMP} $\downarrow$ & \textbf{mIoU (\%)} $\uparrow$ & \textbf{NMP} $\downarrow$ \\
    \midrule\midrule
    I-$512$, V-$0.2$ & $60.5$ & $70.6$ & $291$ & $48.2$ & $985$ \\
    \rowcolor{lora_orange!10} I-$768$, V-$0.15$ & $\mathbf{69.8}$ & $\mathbf{80.1}$ & $\mathbf{280}$ & $\mathbf{55.7}$ & $\mathbf{582}$ \\
    \bottomrule
    \end{tabular}
  \label{tab:res}
  \vspace{-3mm}
\end{table}
\noindent \textbf{Ablation on Input Resolution.}
Next, we examine the role of input resolution in the performance of promotable segmentation. 
Compared to the baseline setting of the image resolution $512\times512$ and the resolution of $0.2$ m voxels, increasing the resolution to $768\times768$ for the images and $0.15$ m for the voxels results in a notable performance gain, as presented in \cref{tab:res}. This demonstrates the importance of high-resolution input in dense prediction tasks, where finer spatial details contribute to more accurate segmentation.

\begin{table}[t]
  \centering
  \caption{
    Ablation study on ego-motion in memory attention. 
    }
    \vspace{-0.3cm}
    \scriptsize
    \renewcommand\tabcolsep{1.2pt}
    \renewcommand\arraystretch{1.1}
    \begin{tabular}{lccccc}
    \toprule
    \multirow{2}{*}{\textbf{MCMA}} & \multicolumn{3}{c}{\textbf{Image}} & \multicolumn{2}{c}{\textbf{LiDAR}} \\
    \cmidrule(lr){2-4} \cmidrule(lr){5-6}
    & \textbf{mIoU (\%)} $\uparrow$ & {$\mathbf{\mathcal{J\&F}}$ $\uparrow$ (\%)} & \textbf{NMP} $\downarrow$ & \textbf{mIoU (\%)} $\uparrow$ & \textbf{NMP} $\downarrow$ \\
    \midrule\midrule
    \textit{w/o.} ego-motion & $69.7$ & $\mathbf{80.3}$ & $298$ & $52.2$ & $746$ \\
    \rowcolor{lora_orange!10} \textit{w.} ego-motion & $\mathbf{69.8}$ & $80.1$ & $\mathbf{280}$ & $\mathbf{55.7}$ &  $\mathbf{582}$\\
    \bottomrule
    \end{tabular}
  \label{tab:ego}
  \vspace{-5mm}
\end{table}
\noindent \textbf{Ablation on Ego-motion in Memory Attention.}
Finally, we perform an ablation study on the incorporation of ego-motion in Motion-aware Cross-modal Memory Attention to assess its contribution to temporal feature fusion and object tracking. As presented in \cref{tab:ego}, ego-motion compensation significantly reduces tracking inconsistencies in stream segmentation, particularly for LiDAR, where NMP decreases from $746$ to $592$, indicating improved temporal stability. Furthermore, its integration leads to a notable improvement in mIoU, highlighting the importance of ego-motion in enhancing segmentation accuracy over long sequences.
\section{Conclusion}
In this paper, SAM4D is introduced as a multi-modal and temporal model for promptable segmentation across camera and LiDAR streams. 
Our contributions span \textbf{task} (PMS), \textbf{model} (SAM4D), and \textbf{data} (Waymo-4DSeg).
With extensive experiments, SAM4D advances 4D scene understanding, improving segmentation consistency, efficiency, and adaptability in various autonomous driving scenarios. 
We believe that the insights of the SAM4D model into multimodal prompting and 4D perception will significantly reduce annotation costs, enabling high-quality scalable 2D-3D joint labeling for large-scale datasets.

\section*{Acknowledgments}

We thank the reviewers for the valuable discussions and our colleagues for preparing the user-interaction system. This research was supported by the Zhejiang Provincial Natural Science Foundation of China under Grant No. LD24F030001.

{
    \small
    \bibliographystyle{ieeenat_fullname}
    \bibliography{main}
}

\appendix
\maketitlesupplementary

\setcounter{figure}{0}
\setcounter{table}{0}
\renewcommand{\thefigure}{A\arabic{figure}}
\renewcommand{\thetable}{A\arabic{table}}

In this document, we further provide the following materials to support the findings and conclusions drawn in the main body of this paper.

\begin{itemize}
    \item Section~\ref{sec:supp_model}: Model and training details;
    \item Section~\ref{sec:supp_dataset}: Details on data engine and dataset;
    \item Section~\ref{sec:supp_exps}: Details on the experimental settings;
    \item Section~\ref{sec:supp_limit}: Discussions on limitation and future work;
    \item Section~\ref{sec:supp_public}: Acknowledgements to public resources.
\end{itemize}

\section{Model and Training Details}
\label{sec:supp_model} 
\subsection{Model Architecture}
\noindent \textbf{Image Encoder.} The image encoder in SAM4D follows the same architecture as SAM2~\cite{ravi2024sam}, but given the smaller scale of our dataset compared to SAM2, we adopt the Hiera-S variant~\cite{ryali2023hiera} as the default configuration to balance performance and efficiency.

\noindent \textbf{LiDAR Encoder.}
We adopt MinkUNet~\cite{choy20194d}, implemented with TorchSparse~\cite{tang2022torchsparse,tangandyang2023torchsparse++}, as the LiDAR encoder. Inspired by ResNet~\cite{he2016deep}, we define Mink34 and Mink50 as backbone structures, with Mink34 as the default choice. 
The encoder downsamples the input to stride $32$, with feature dimensions $[32, 32, 64, 128, 256]$, and then upsamples to stride $4$, extracting voxel features at strides $16$, $8$, and $4$. The stride 16 features are primarily used by the memory module and mask decoder, while the stride $8$ and $4$ features assist the mask decoder in recovering high-resolution segmentation details, similar to SAM2.
To improve the generalization across various datasets, we exclude the original xyz coordinates and do not use intensity and elongation features provided by Waymo Open Dataset~\cite{sun2020scalability}. Instead, we assign a binary occupancy value to occupied voxels, ensuring the LiDAR encoder remains dataset-agnostic.

\noindent \textbf{Mask Decoder.}
Sparse prompt tokens from image and LiDAR are concatenated and used as queries for mask prediction, with a shared Transformer module across both modalities. The query token configuration follows SAM2, consisting of mask queries, sparse prompt tokens, IoU tokens, and object pointers stored in memory. 
This design enables efficient cross-modal segmentation while ensuring consistency between image and LiDAR-based queries.

\noindent \textbf{Model Parameters.}
SAM4D comprises $119.88$M parameters, distributed across its core components. The image encoder has $34.32$M parameters, while the LiDAR encoder contains $26.94$M parameters. The memory module, responsible for temporal feature aggregation and cross-modal attention, is the largest component with $53.96$M parameters. The mask decoder, which processes prompt-based queries for segmentation, accounts for$4.66$M parameters. This design balances multimodal fusion, temporal reasoning, and segmentation efficiency.

\begin{table}[t]
\centering
\scriptsize
\renewcommand\tabcolsep{15.0pt}
\renewcommand\arraystretch{1.1}
\begin{tabular}{l|l}
\toprule
\textbf{Config} & \textbf{Value} \\
\midrule\midrule
data & Waymo-4DSeg \\
steps & $\sim$$44$k \\
resolution & Camera: $768\times768$, LiDAR: $0.15$ m \\
precision & bfloat16 \\
optimizer & AdamW \\
optimizer momentum & $\beta_1=0.9, \beta_2=0.999$ \\ 
gradient clipping & type: $\ell_2$, max: $0.1$ \\
weight decay & $0.1$ \\
learning rate (lr) & OneCycleLR, init: $5e^{-6}$, max: $5e^{-5}$, \\
& anneal strategy: cos, pct\_start: $0.4$ \\
warmup & linear, $7.5$k iters \\ 
layer-wise decay & $0.9$ \\
image augmentation & hflip, resize to $768\times768$ (square) \\
video augmentation & hflip, affine (deg: $25$), colorjitter, \\
& grayscale, per-frame colorjitter, \\
& mosaic-$2\times2$ \\
LiDAR augmentation & rotation-Z (deg: $45$), hflip, vflip \\
drop path & $0.2$ \\
mask losses (weight) & Focal ($20$), Dice ($1$) \\
IoU loss (weight) & $\ell_1$ ($1$) \\
occlusion loss (weight) & Cross-entropy ($1$) \\
global attn. blocks & $12-16-20$ \\
\bottomrule
\end{tabular}
\caption{Hyperparameters for SAM4D full training.}
\label{tab:sam4d-training-hparam}
\end{table}

\subsection{Training Details}
Without loss of generality, we use the front-view camera and LiDAR from the Waymo dataset to validate the feasibility of the proposed solution, which can later be extended to multi-camera and LiDAR setups. Since our data is constructed through 4D reconstruction using a $5$-camera and LiDAR system, we cannot guarantee that objects appear in both modalities in every frame. Therefore, during the training process, to enable parallel imitation of interaction logic for multiple targets, the following rules are applied when selecting targets for each step in the data pipeline: there is a $0.5$ probability that the target exists in both modalities, a $0.25$ probability that it appears only in the camera, and a $0.25$ probability that it appears only in the LiDAR. During training, the iterative modification logic for mimicking targets is as follows: if the target belongs to both modalities, each prompt randomly selects one modality; if the target belongs to only one modality, the modality in which the target appears is chosen for the prompt.

SAM4D is trained on Waymo-4DSeg for $44$k steps with a $768\times768$ image resolution and a LiDAR voxel size of $0.15$. Specifically, we sample $8$-frame sequences and randomly select up to $2$ frames to receive prompts. During training, corrective clicks are probabilistically sampled based on both ground-truth masks and model predictions. The initial prompts are assigned with probabilities of $0.5$ for ground-truth masks, $0.25$ for points, and $0.25$ for bounding boxes, respectively. The loss consists of a combination of focal loss and dice loss for mask prediction and mean absolute error (MAE) loss for IoU prediction. If an object is missing in a given modality, we do not apply supervision to the prediction of that modality. AdamW with OneCycleLR are utilized to optimize the network. Image and video augmentations follow SAM2 (excluding shear), while LiDAR-specific augmentations include Z-axis rotation, hflip, and vflip. Other hyperparameters align with SAM2, which are provided in \cref{tab:sam4d-training-hparam}.

\section{Details on Data Engine and Dataset}
\label{sec:supp_dataset}

\begin{figure*}[t]\centering
\includegraphics[width=1.0\linewidth]{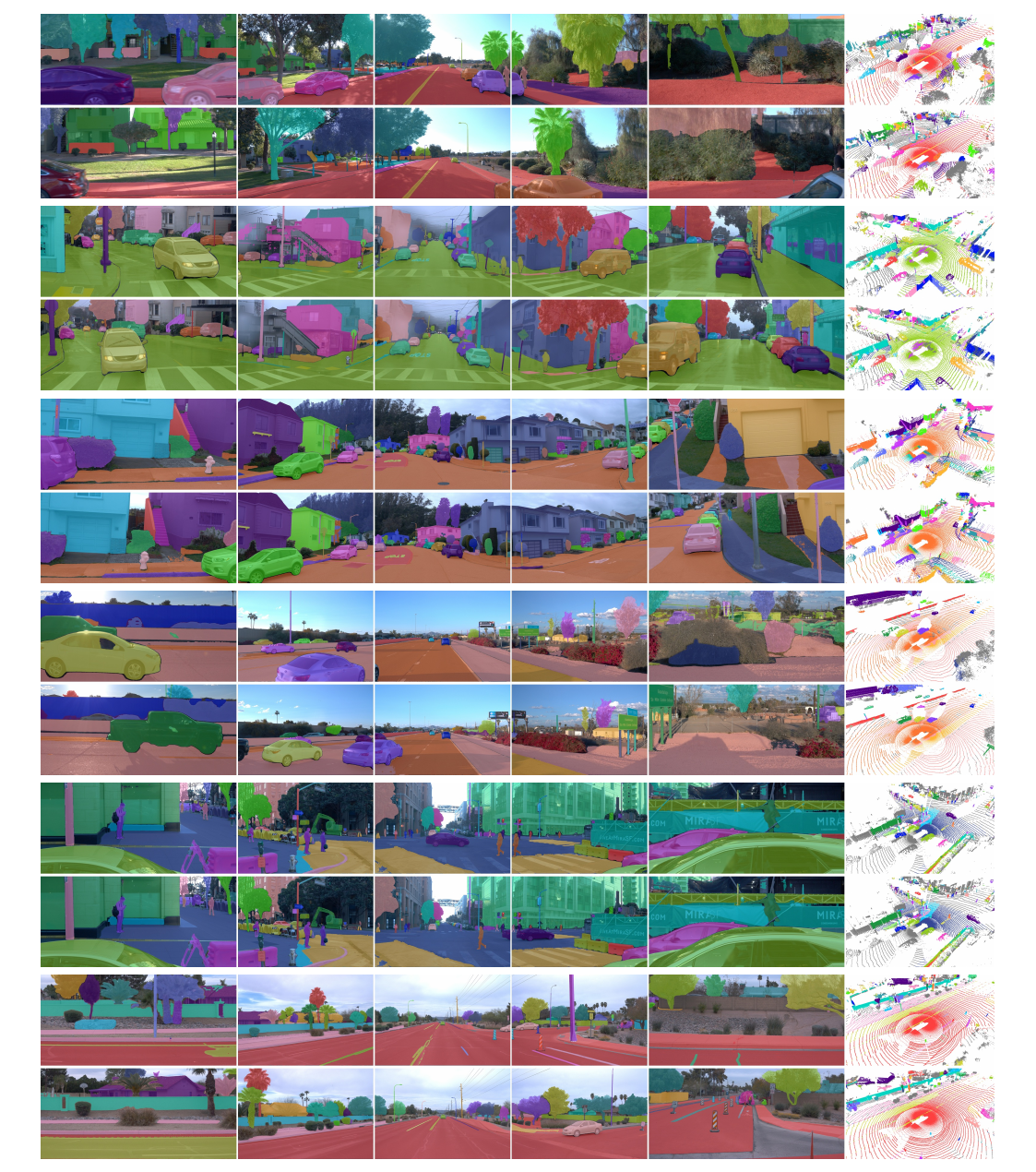}\vspace{-1mm}
\caption{
Examples of the Waymo-4Dseg Dataset: In this figure, we visualize the masklets from selected frames of 6 clips. For each clip, we present two frames in two rows, where each row displays the masklets from the side-left, front-left, front, front-right, and side-right images, from left to right, along with the masklets from the LiDAR frames. Through both horizontal and vertical comparisons, the consistency of the masklets across different video streams, modalities, and frames can be observed.
}
\label{fig:dataset_demo}
\vspace{-1mm}
\end{figure*}

\subsection{Data Engine Details}

In this section, we provide a more detailed description of the implementation details of the three steps in our data engine to supplement Sec. \textcolor{red}{5} in our main paper.

\noindent \textbf{Step 1: Generation of VFM-based image masklets}

In this step, we utilize vision foundation models (VFM)~\cite{liu2024grounding, chen2024sam2} to generate initial annotations including boxes and masks in keyframes firstly. 
In each new keyframe, we redetect scene objects and match them with the propagated masks from the previous keyframe, merging them as the segmentation result of the current keyframe. Newly detected objects, which were not present in the previous masklets, are first propagated backward to the start of the sequence, and then the merged masks (both new and existing objects) are propagated forward to the next keyframe, continuing this process until the end of the sequence.

This iterative approach produces masklets for the entire image sequence, ensuring consistent object categories and instance IDs across time, laying the foundation for LiDAR ground truth generation in subsequent stages.

\noindent \textbf{Step 2: 4D Reconstruction and Ray Casting}

Transferring masklets from video to LiDAR frames requires establishing correspondences between pixels and LiDAR points, which is challenging due to the large number of pixels and 3D points in the sequence. 
To address this challenge, we preemptively perform 4D LiDAR reconstruction using VDBFusion~\cite{vizzo2022vdbfusion}, which generates a more efficient representation of spatial occupancy, since the voxel count depends only on scene size and is independent of the number of frames.

The 4D reconstruction comprises multiple foreground components and a single background component. The background consists of static objects and is maintained as a single instance in the world coordinate system. The foreground includes potentially moving entities such as vehicles, cyclists, and pedestrians, each with its own motion trajectory. We leverage the pre-annotated 3D bounding boxes of these foreground objects to obtain their relative poses in each frame with respect to the world, and subsequently perform individual 4D reconstruction within each object's body coordinate system. This allows the voxels occupied by the foreground objects to remain unchanged even as they move, with only the overall position shifting.

Following this, we generate a dense pixel-voxel mapping table via ray casting. We first compute image poses in the world coordinate system by solving the PnP(Perspective-n-Point) problems, based on the given single-frame point cloud and pixel correspondences. Then, for each image, we construct multiple rays starting from the camera position and ending at the center points of the voxels within the viewing frustum. Each ray falls into an image pixel, and we match the pixel with the voxel that the ray intersects together to build pixel-voxel the mapping table.

\begin{figure*}[t]\centering
\includegraphics[width=1.0\linewidth]{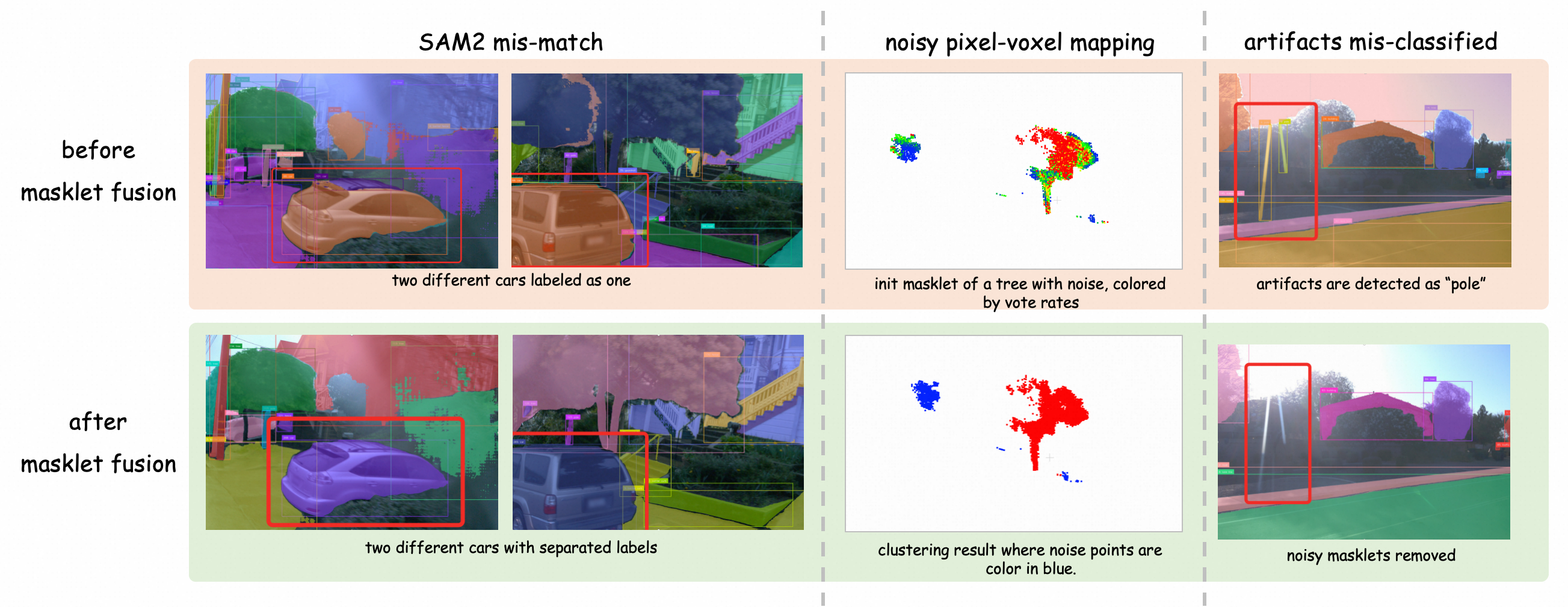}\vspace{-3mm}
\caption{
This figure illustrates three representative problems that may arise before masklet fusion, as well as how our fusion process addresses these issues.
}
\label{fig:fusion_benefits}
\vspace{-4mm}
\end{figure*}

\noindent \textbf{Step 3: Cross-Modal Masklet Fusion}

By querying the pixel-voxel mapping table established in Step 2, we can identify the voxels corresponding to the pixels masked in Step 1, thereby transferring the mask to the voxels.  
In an ideal scenario, SAM2 ensures consistency of masklets between frames in the video, and we can directly merge the voxel masks from the video stream to obtain an accumulated voxel masklet.

However, we found that both the video masklets and the mapping table are often noisy. A common issue is that SAM2 mis-matches objects, confusing two similar objects appearing in close positions across different frames as the same object. This issue occurs for both background and foreground objects. Additionally, because the pixel-to-voxel correspondence is not always accurate and the edges of 2D masks on images are not perfectly precise, the resulting masks projected onto the voxels are prone to contain noise. Finally, the image segmentation model occasionally misclassifies artifacts such as light spots or other visual anomalies in images as actual objects.

To mitigate these problems, we implemented a clustering approach for noise filtering in the voxel masklets. We employed the DBSCAN algorithm to cluster voxels based on their BEV positions and selected the cluster with the highest average quality as the main cluster, filtering out the rest as noise. Assuming that voxels associated with a single object are adjacent in BEV space, we utilized the DBSCAN algorithm to cluster the voxels based on their BEV positions. We also counted the frequency with which each voxel was mapped to the current object and computed the vote rate as the ratio of this frequency to the total observations in the current image sequence. Ultimately, we selected the cluster with the highest average vote rate as the main cluster, and leave out rest as noise. Fig ~\ref{fig:fusion_benefits} provides examples of the issues mentioned above and demonstrates the effectiveness of our filter in addressing these problems.

After filtering, we expect significant overlap between voxel masklets from different videos corresponding to the same object. Overlaps between masklets from two videos were assessed, leading to the merging of those with substantial overlap into a single unified voxel masklet. Finally, we created a mapping table between points from LiDAR frames and voxels based on their 3D spatial distances, facilitating the transfer of the final voxel masklet to the LiDAR frames.

We evaluated the quality of the unified voxel masklets using cross-modal IoU. Assuming that a masklet is visible for image \textit{i}, we calculated the IoU between the voxels mapped by masklet in image \textit{i} and the visible part of the unified voxel masklet. The average IoU across all images represents one masklet's overall score. The mean score of the masklets in our dataset is 0.56, with a 10th percentile of 0.24. Throughout this process, human annotators play a crucial role in adjusting the parameters based on mask quality and conducting frame-by-frame verification of the final labels in both the image and LiDAR.

\subsection{Dataset Statistical Information}

\begin{figure}[t]\centering
\includegraphics[width=0.99\linewidth]{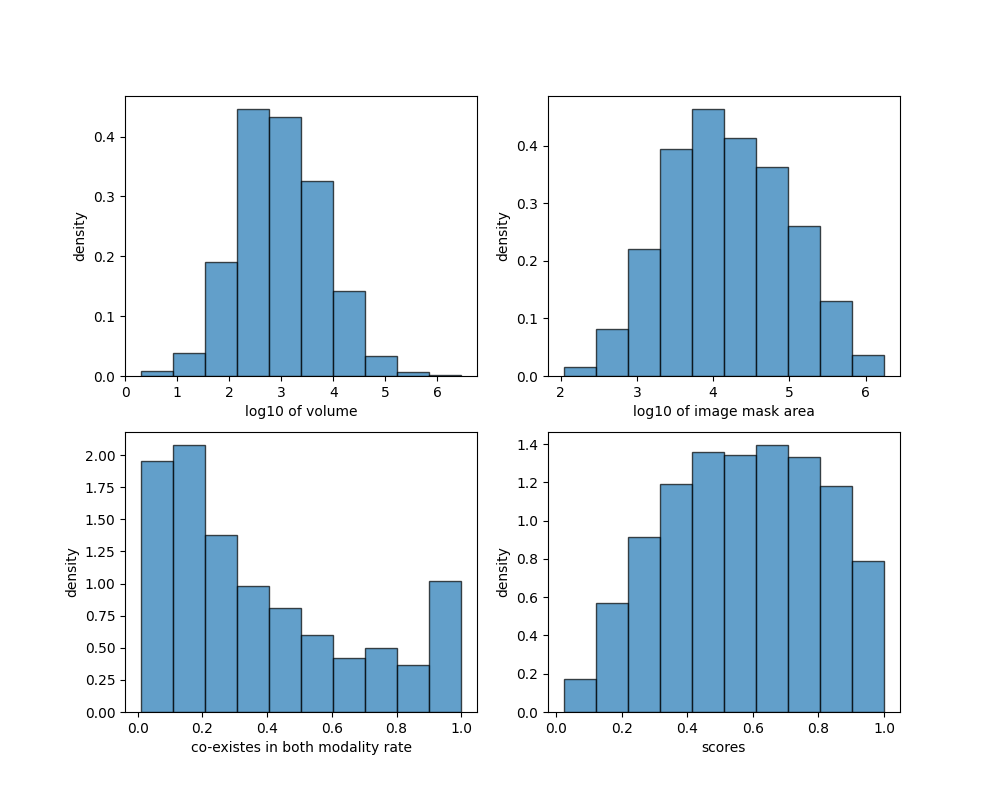}\vspace{-1mm}
\caption{Statistical information of masklets in Waymo-4DSeg dataset}
\label{fig:dataset_stat}
\vspace{-4mm}
\end{figure}

Here, we provide additional information about the dataset and details about our data engine.
In Figure ~\ref{fig:dataset_stat}, we provide detailed statistical information about the masklets in our dataset, including volume distribution, area distribution, the proportion of frames in which cross-modal masklets co-occur, and score distribution.
The distributions of volume and area reflect the diversity and richness of the annotated objects in our dataset. Additionally, we calculate the proportion of frames in which cross-modal masklets are present in both modalities, a metric of interest to users. Common scenarios include objects exiting the video frame as the vehicle moves forward while still being detectable by LiDAR, or objects being beyond the scanning range of LiDAR but still visible in the video. Such cases require special handling during model prompting and training, which is why we provide the proportion of frames where cross-modal masklets coexist.
We also present the score distribution of the masklets. Although denoising has been applied during processing, a small number of low-quality masklets remain. Fortunately, we can quantify their quality through scores and filter them out during experimentation. In Figure ~\ref{fig:dataset_demo}, we present the visualized cross-modal masklets from several clips.

\section{Details on the Experimental Settings}
\label{sec:supp_exps}
\subsection{About the Training and Validation Data.} 
As mentioned above, when the data engine generates pseudo labels, it creates voxels in 4D for each object, allowing us to obtain the volume of the object in space. Additionally, multi-modal consistency checks are performed to calculate the IoU between each frame's image and point cloud. The average IoU over the sequence serves as a reference for the quality of the pseudo-labels, which we refer to as the "score." During SAM4D training, targets with a volume greater than 10 and a score greater than 0.3 are used. For testing, to further ensure the reliability of the ground truth, the volume filtering threshold is increased to 50, and the score threshold is raised to 0.5. Furthermore, there is currently significant ambiguity in the pseudo-labels for ground regions. To ensure better convergence of the LiDAR branch, we temporarily exclude instances near the ground during both training and evaluation. Despite these settings, the number of targets evaluated in each sequence still exceeds 100, resulting in slow evaluation speeds. To accelerate the evaluation, we filter and evaluate only those objects that appear in at least one frame in both the front-view camera and LiDAR.

\subsection{About the Generalization Experiments.} 
The generalization experiments are conducted on nuScenes dataset~\cite{caesar2020nuscenes} with nuInsSeg~\cite{li2023lwsis}.
The nuInsSeg dataset~\cite{li2023lwsis}, built on nuScenes~\cite{caesar2020nuscenes}, provides 2D instance segmentation annotations for foreground objects, with instance IDs corresponding to 3D point cloud segmentation labels.

\section{Limitations and Further Discussions}
\label{sec:supp_limit}
\subsection{Limitations}
While SAM4D effectively integrates multimodal and temporal segmentation, the domain gap across LiDAR sensors remains a challenge, as variations in sensor configurations and point cloud density limit generalization compared to images. Moreover, the spatial representation on point clouds is inherently constrained by single-frame sparsity, occlusions, and blind spots, which may hinder object completeness in certain scenarios. Additionally, while Waymo-4DSeg provides high-quality multimodal labels, the size of the data set can be expanded to cover a broader range of driving conditions, weather variations, and rare long-tail scenarios.
Increasing data set diversity would improve the generalizability of the model, particularly in corner cases where data sparsity remains a challenge.

\subsection{Future Work}
Currently, SAM4D is trained on pseudo-labels generated by an automated data engine. Although the data labels have undergone multi-modal consistency verification, ambiguities and inaccuracies still persist. Future work will focus on improving SAM4D’s data strategy, model adaptability, and scalability. 
To enhance label quality, we plan to expand dataset scale using our automated data engine and integrate human-annotated subsets for fine-tuning. A confidence-based filtering mechanism will further refine pseudo-labels iteratively. Additionally, extending SAM4D to incorporate natural language descriptions will enable multimodal segmentation conditioned on text, leveraging LLMs for semantic guidance to reduce reliance on human annotations. Exploring weakly supervised and self-supervised learning will further enhance adaptability while minimizing manual labeling. Beyond data efficiency, improving memory attention and computational efficiency will enable scaling to multi-camera and multi-sensor systems, enhancing 4D spatiotemporal perception in complex environments.

\section{License and Consent with Public Resources}
\label{sec:supp_public}
\subsection{Public Datasets}

We utilize the Waymo Open Dataset~\cite{caesar2020nuscenes} to construct our Waymo-4DSeg dataset. nuScenes~\cite{caesar2020nuscenes} and the corresponding nuInstSeg~\cite{li2023lwsis} are adopted to further evaluate our model:
\begin{itemize}
    \item Waymo Open Dataset\footnote{\url{https://waymo.com/open}.} \dotfill  Waymo Dataset License
    \item nuScenes\footnote{\url{https://www.nuscenes.org/nuscenes}.} \dotfill CC BY-NC-SA 4.0
    \item nuScenes-devkit\footnote{\url{https://github.com/nutonomy/nuscenes-devkit}.} \dotfill Apache License 2.0
    \item nuInsSeg\footnote{\url{https://github.com/Serenos/nuInsSeg}.} \dotfill MIT License
    
\end{itemize}

\subsection{Public Implementation}
We leverage publicly available pre-trained models and source codes to investigate the promptable segmentation in the multimodal domain:
\begin{itemize}
     \item SAM2\footnote{\url{https://github.com/facebookresearch/sam2}.} \dotfill Apache License 2.0
     \item TorchSparse\footnote{\url{https://github.com/mit-han-lab/torchsparse}.} \dotfill MIT License
     \item GroundingDINO\footnote{\url{https://github.com/IDEA-Research/GroundingDINO}.} \dotfill Apache License 2.0
     \item Grounded-SAM-2\footnote{\url{https://github.com/IDEA-Research/Grounded-SAM-2}.} \dotfill Apache License 2.0
     \item VDBFusion\footnote{\url{https://github.com/PRBonn/vdbfusion}.} \dotfill MIT License
     \item Mask-Propagation\footnote{\url{https://github.com/hkchengrex/Mask-Propagation}.} \dotfill MIT License
     
\end{itemize}

\end{document}